\documentclass[nojss]{jss}

\usepackage{orcidlink,thumbpdf,lmodern}

\usepackage[utf8]{inputenc}

\author{
Pablo Morala\\Universidad Carlos III de Madrid \And J. Alexandra
Cifuentes\\Universidad Pontificia Comillas \AND Rosa E.
Lillo\\Universidad Carlos III de Madrid \And Iñaki Ucar\\Universidad
Carlos III de Madrid
}
\title{\pkg{nn2poly}: An \proglang{R} Package for Converting Neural
Networks into Interpretable Polynomials}

\Plainauthor{Pablo Morala, J. Alexandra Cifuentes, Rosa E. Lillo, Iñaki
Ucar}
\Plaintitle{nn2poly: An R Package for Converting Neural Networks into
Interpretable Polynomials}

\Abstract{
The \pkg{nn2poly} package provides the implementation in \proglang{R} of
the NN2Poly method to explain and interpret feed-forward neural networks
by means of polynomial representations that predict in an equivalent
manner as the original network. Through the obtained polynomial
coefficients, the effect and importance of each variable and their
interactions on the output can be represented. This capabiltiy of
capturing interactions is a key aspect usually missing from most
Explainable Artificial Intelligence (XAI) methods, specially if they
rely on expensive computations that can be amplified when used on large
neural networks. The package provides integration with the main deep
learning framework packages in \proglang{R} (\pkg{tensorflow} and
\pkg{torch}), allowing an user-friendly application of the NN2Poly
algorithm. Furthermore, \pkg{nn2poly} provides implementation of the
required weight constraints to be used during the network training in
those same frameworks. Other neural networks packages can also be used
by including their weights in \code{list} format. Polynomials obtained
with \pkg{nn2poly} can also be used to predict with new data or be
visualized through its own plot method. Simulations are provided
exemplifying the usage of the package alongside with a comparison with
other approaches available in \proglang{R} to interpret neural networks.
}

\Plainkeywords{R, neural networks, interpretability, XAI, polynomials}

\Address{
    Pablo Morala\\
    Universidad Carlos III de Madrid\\
    Deparment of Statistics and uc3m-santander Big Data Institute.\\
Calle Madrid 135, 28903 Getafe (Madrid), Spain.\\
  E-mail: \href{mailto:pablo.morala@uc3m.es}{\nolinkurl{pablo.morala@uc3m.es}}\\
  
      J. Alexandra Cifuentes\\
    Universidad Pontificia Comillas\\
    Department of Quantitative Methods, Faculty of Economics and
Business Administration\\
and the Institute for Research in Technology (IIT), ICAI School of
Engineering.\\
Calle de Alberto Aguilera 23, 28015 Madrid, Spain\\
  E-mail: \href{mailto:jacifuentes@comillas.edu}{\nolinkurl{jacifuentes@comillas.edu}}\\
  
      Rosa E. Lillo\\
    Universidad Carlos III de Madrid\\
    Deparment of Statistics and uc3m-santander Big Data Institute.\\
Calle Madrid 135, 28903 Getafe (Madrid), Spain.\\
  E-mail: \href{mailto:lillo@est-econ.uc3m.es}{\nolinkurl{lillo@est-econ.uc3m.es}}\\
  
      Iñaki Ucar\\
    Universidad Carlos III de Madrid\\
    Deparment of Statistics and uc3m-santander Big Data Institute.\\
Calle Madrid 135, 28903 Getafe (Madrid), Spain.\\
  E-mail: \href{mailto:inaki.ucar@uc3m.es}{\nolinkurl{inaki.ucar@uc3m.es}}\\
  
  }

\providecommand{\tightlist}{%
  \setlength{\itemsep}{0pt}\setlength{\parskip}{0pt}}

\usepackage{amsmath} \usepackage{longtable} \usepackage{booktabs} \usepackage{subfig} \usepackage{algpseudocode} \usepackage{algorithm}   \usepackage{mathtools} \usepackage{amsthm} \newtheorem{theorem}{Theorem}[section] \newtheorem{lemma}[theorem]{Lemma}

\begin{document}

\hypertarget{introduction}{%
\section{Introduction}\label{introduction}}

\label{section:introduction}

Machine Learning (ML) and Artificial Intelligence (AI) have based their
recent advances and success mainly on the use of neural networks,
specially with the development of deep learning
\citep{lecunDeepLearning2015}. Despite their widespread adoption in
almost every field, several concerns exist still about their inner
workings, specifically about their opaque nature, often considered black
boxes
\citep{benitezAreArtificialNeural1997, shwartz-zivOpeningBlackBox2017}.
Significant efforts have been aimed towards increasing the
explainability or interpretability of NNs, opening research fields such
as eXplainable Artificial Intelligence (XAI).

Among the solutions proposed to address the black-box nature of AI, we
can differentiate between model agnostic methods (such as SHAP
\citep{lundbergUnifiedApproachInterpreting2017} or LIME
\citep{ribeiroWhyShouldTrust2016}) and other solutions tailored for an
specific model, which in the neural networks case can be found in
examples such as Layer-wise Relevance Propagation (LRP)
\citep{bachPixelWiseExplanationsNonLinear2015}, Gradient based methods
\citep{baehrensHowExplainIndividual2010, simonyanDeepConvolutionalNetworks2014}
with variants such as SmoothGrad
\citep{smilkovSmoothGradRemovingNoise2017}, DeepLift
\citep{shrikumarLearningImportantFeatures2019}, or sensitivity based
analysis methods like the one implemented in \pkg{NeuralSens}
\citep{pizarrosoNeuralSensSensitivityAnalysis2022}. In this context,
there have been several proposals trying to bring together neural
networks and polynomial models, as the polynomials are an inherently
interpretable models due to the explanation of their coefficients. An
initial proposed approach was to explore the integration of polynomial
functions with neural networks, drawing inspiration from classical
statistical techniques
\citep{chenAnalysisDesignMultiLayer1998, chenBackpropagationRepresentationTheorem1990, chenConventionalModelingMultilayer1993}.
This idea was based in the universal approximation theorems of the 1990s
\citep{cybenkoApproximationSuperpositionsSigmoidal1989, hornikMultilayerFeedforwardNetworks1989, hornikApproximationCapabilitiesMultilayer1991},
using polynomial basis functions to establish relationships with
multilayer perceptrons (MLPs), offering an alternative framework for
model representation and analysis. This methodology relied on solving
optimization problems via a Least-Mean-Square approach to determine
polynomial coefficients.

In recent years, with the advances in deep learning, there has been a
renewed interest in the union between neural networks and polynomial
models, with methods such as Deep Polynomial Neural Networks (or
\(\Pi\)-networks) \citep{chrysosDeepPolynomialNeural2022}. These models
use high-order polynomials to capture complex interactions, which are
trained by representing interactions via high-order tensors and
employing factor sharing techniques. This approach allows these models
to reduce the computational cost associated with modeling all possible
interactions \citep{rendleFactorizationMachines2010}. Additionally,
other architectures like quadratic networks have emerged, increasing
network expressiveness through quadratic functions instead of
traditional inner products within neurons
\citep{fanExpressivityTrainabilityQuadratic2023}. In general, these
advancements showcase the increasing interest on polynomial models as
viable alternatives to traditional neural networks, particularly in
tasks requiring interpretability and feature interaction modeling.

Following this trend of trying to benefit from the inherently
interpretable nature of polynomials, the NN2poly method
\citep{moralaNN2PolyPolynomialRepresentation2023a} proposes an algorithm
to transform pre-existing MLPs, i.e., already trained networks, into
polynomial representations. This method extends to deeper layers a
previous approach that was limited to single hidden layer MLPs
\citep{moralaMathematicalFrameworkInform2021}. Unlike other polynomial
approaches that focus on training specialized architectures with
polynomial outputs, NN2Poly enables the interpretation of existing
neural networks through polynomial approximations of their internal
parameters. Layer by layer, the polynomials that represent the network
are built iteratively, using the internal weights and activation
functions approximated by Taylor expansion. It is important to note that
for these approximations to work, certain weight constraints need to be
imposed during training. This method is particularly suited for tabular
data applications, where the statistical interpretation of polynomial
coefficients are directly applicable. Furthermore, the obtained
polynomials provide interpretation, through their coefficients, to
variable interactions instead of single variables, a key difference with
many of the most adopted interpretability approaches.

This article presents the \pkg{nn2poly} package \citep{R-nn2poly} for
\proglang{R} \citep{rcoreteamLanguageEnvironmentStatistical2023a}, which
allows the application of the NN2Poly method to MLP neural networks and
therefore explain their predictions using an equivalent polynomial, with
interpretable coefficients both for the single variables and their
interactions. The package supports the two current main deep learning
frameworks, \pkg{Torch} and \pkg{Tensorflow}, while neural networks from
other arbitrary frameworks can easily be used by extracting their
weights and activation functions. Additionally, the package provides
functions to impose needed constraints during the neural network
training in \pkg{Torch} and \pkg{Tensorflow}.

The rest of the article is structured as follows. The remaining part of
Section \ref{section:introduction} provides an overview on
interpretability software and neural network training software in
\proglang{R}. Then, Section \ref{section:theoretical} presents the
fundamental concepts and theory behind the NN2Poly method. Section
\ref{section:pkg_struc} explains in detail the \pkg{nn2poly} package
functions and structure with regression examples based on polynomial
data. Section \ref{section:classification} shows the method employed on
a classification example and Section \ref{section:comparison} compares
the obtained results on the polynomial examples when explained through
other alternative packages. Finally, Section \ref{section:conc}
summarizes the contributions of this work.

\hypertarget{related-interpretability-software}{%
\subsection{Related interpretability
software}\label{related-interpretability-software}}

\label{section:related_interpret}

Regarding related software, there exists a broad offer of
interpretability or XAI packages in \proglang{R} with general model
agnostic interpretability options, and also some with neural networks
specific options. The following list provides an overview of the current
options available in \proglang{R} for these matters:

\begin{itemize}
\tightlist
\item
  \textbf{SHAP} based interpretability: SHappley Additive exPlanations
  (SHAP) \citep{lundbergUnifiedApproachInterpreting2017} has been
  established as one of the main methods in interpretable and
  explainable machine learning, along with its many extensions. This
  method quantifies the contribution of each feature to the final model
  based on cooperative game theory concepts. It has an official
  implementation by the authors in \proglang{Python}. In \proglang{R},
  there are several implementations, such as \pkg{shapper}
  \citep{R-shapper} which is a wrapper of the \proglang{Python}
  implementation, or \pkg{fastshap} \citep{R-fastshap}, \pkg{explainer}
  \citep{explainer2024}, \pkg{shapr} \citep{R-shapr} and \pkg{shapley}
  \citep{R-shapley} that provide alternative implementations with
  different improvements. There are also several model specific SHAP
  versions, specially focused on tree based model, but they are not
  applicable to neural networks. Furthermore, there are also
  visualization specific packages such as \pkg{shapviz}
  \citep{R-shapviz}, which allows the creation of importance plots from
  different interpretability results.
\item
  \textbf{LIME} based interpretability: Local Interpretable
  Model-agnostic Explanations (LIME) \citep{ribeiroWhyShouldTrust2016}
  is another method widely used in the XAI community, which provides
  explanations by learning an interpretable model locally around the
  desired prediction. This approach is model agnostic also, and its
  original implementation is provided in \proglang{Python}. However,
  there are also implementations in \proglang{R} such as \pkg{lime}
  \citep{R-lime} which is a port of the original implementation or
  \pkg{live} \citep{RJ-2018-072} and \pkg{localModel}
  \citep{R-localModel} which are alternative implementations.
\item
  \textbf{Other model agnostic methods}: Besides SHAP and LIME, there
  are other proposals for model agnostic methods that can also be
  employed on neural networks. Some of them, but not all, are: \pkg{ale}
  \citep{ale2023} implementing ALE plots, \pkg{iml} \citep{iml2018} and
  \pkg{distillML} \citep{R-distillML} which both implement several
  methods like PDP plots, ALE plots, surrogate models or even SHAP
  values in the former, \pkg{breakDown} \citep{RJ-2018-072} implements
  another method that decomposes predictions into parts that can be
  attributed to specific features, and \pkg{DALEX} \citep{DALEX2018}
  implements an ecosystem with several levels of explanations, some of
  them also based on SHAP.
\item
  \textbf{Neural Network specific} interpretability:

  \begin{itemize}
  \tightlist
  \item
    \pkg{NeuralSens} \citep{gonzalezNeuralSensSensitivityAnalysis2024}:
    This package provides sensitivity analysis of neural networks using
    a partial derivatives method. This method is implemented
    specifically for MLPs (same as NN2Poly), where partial derivatives
    of the output with respect to the input are computed. Several plots
    and graphical interpretations of the results are provided.
  \item
    \pkg{innsight} \citep{koenenInnsightGetInsights2023}: This package
    implements several XAI techniques for neural networks, such as LRP,
    DeepLift or gradient-based methods, among other model agnostic
    methods applicable to neuronal networks like SHAP or LIME.
  \item
    \pkg{NeuralNetTools}
    \citep{beckNeuralNetToolsVisualizationAnalysis2018}: This package
    provides visualization of neural network weights and feature
    importance through the Olden
    \citep{oldenAccurateComparisonMethods2004} and Garson
    \citep{garsonInterpretingNeuralnetworkConnection1991} methods.
  \end{itemize}
\end{itemize}

This overview covers the majority of applicable methods to
interpretability of neural networks in \proglang{R} either through
specific methods or model agnostic methods. Additionally, there are many
specific packages for other models, specially for tree based models or
XGBoost, such as \pkg{treeshap} \citep{R-treeshap} or
\pkg{randomForestExplainer} \citep{R-randomForestExplainer}.
Furthermore, some packages implement inherently interpretable models
such as \pkg{neuralGAM} \citep{R-neuralGAM}, a DL framework based on
Generalized Additive Models, and \pkg{interpret} \citep{R-interpret},
which provides an implementation of Explainable Boosting Machines in
\proglang{R} based on its \proglang{Python} implementation.

However, to the authors knowledge, the specific options for neural
networks interpretability do not include feature interactions. While
some of the general XAI techniques have some proposed extensions to
include interactions (such as SHAP), they are usually quite
computationally expensive,
\citep{sundararajanShapleyTaylorInteraction2020, bordtShapleyValuesGeneralized2023, tsaiFaithShapFaithfulShapley2023}.

\hypertarget{related-neural-networks-training-software}{%
\subsection{Related neural networks training
software}\label{related-neural-networks-training-software}}

\label{section:related_network_training}

Another key aspect of the package is the support provided for different
options in the neural network training and even the possibility of
imposing the custom constraints during training that allow an accurate
approximation by NN2Poly. In this context, there is a significant
difference between frameworks that allow the training of single hidden
layers MLP and frameworks that have full capability of building deep
learning models with several hidden layers. The two currently
\pkg{nn2poly} supported models are the currently most used deep learning
frameworks:

\begin{itemize}
\item
  \pkg{Tensorflow} \citep{abadiTensorFlowLargeScaleMachine2015} is an
  end-to-end machine learning platform. It has been one of the most
  widely used deep learning frameworks in recent years, both in research
  and industry, and it is supported and developed by Google. Their
  original implementation is built in \proglang{Python}, but there is
  also an implementation in the \proglang{R} package \pkg{tensorflow}
  \citep{allaireTensorflowInterfaceTensorFlow2024}, which under the hood
  uses \proglang{Python}. Usually, \pkg{Tensorflow} is used together
  with the deep learning high level API \proglang{Keras}
  \citep{cholletKeras2015}, also originally developed in
  \proglang{Python}. However, it also has an implementation in the
  \proglang{R} package \pkg{keras}
  \citep{kalinowskiKerasInterfaceKeras2023} that allows to work with
  this API from R, using \proglang{Python} again under the hood.
  \pkg{nn2poly} supports the usage of \pkg{tensorflow} models through
  \pkg{keras}.
\item
  \pkg{Torch} \citep{collobertTorch7MatlablikeEnvironment2011} was a
  scientific computing and machine learning framework developed in 2011,
  which was the precursor of \pkg{PyTorch}
  \citep{paszkePyTorchImperativeStyle2019}. \pkg{PyTorch} has also been
  established as the main deep learning framework together with
  \pkg{Tensorflow}, both in research and industry. An implementation in
  \proglang{R} is provided in package \pkg{torch}
  \citep{falbelTorchTensorsNeural2023}, which is the core package of the
  torch ecosystem in \proglang{R} It provides GPU accelerated
  computations and general neural network abstractions for deep learning
  models inspired on \pkg{PyTorch}. However, differing from the
  \pkg{tensorflow} package explained previously, in this case
  \proglang{Python} is not needed to use the \pkg{torch} package, as it
  is natively implemented in \proglang{R} In a similar manner as
  \pkg{keras}, the torch ecosystem contains package \pkg{luz}
  \citep{falbelLuzHigherLevel2023} a higher level API implementation for
  easier model definition and training. There exists another framework
  based on \pkg{torch} in the form of the package \pkg{cito}
  \citep{cito2023}, which uses a syntax inspired by standard
  \proglang{R} and other statistical packages. Currently, \pkg{nn2poly}
  supports the usage of \pkg{torch} models through \pkg{luz}. Support
  for \pkg{cito} is expected to be added in future versions.
\end{itemize}

There are also other options in \proglang{R} that are not currently
supported by \pkg{nn2poly}, but may be included in the future. However,
any model can be used as input to \pkg{nn2poly} as long as the weight
matrices can be extracted. Furthermore, to be able to employ
\pkg{nn2poly} accurately, it is advisable that custom constraints could
be implemented in the chosen framework, condition that not all the
available packages fulfill. Some examples of these other possible neural
network training current options are \pkg{neuralnet}
\citep{R-neuralnet}, which supports MLP \pkg{ANN2} \citep{R-ANN2} and
\pkg{nnlib2Rcpp} \citep{nnlib2Rcpp2021}, both based on Cpp, or general
packages for ML model training that include some forms of neural
networks such as \pkg{h2o} \citep{R-h2o} or \pkg{caret}
\citep{caret2008}.

\clearpage

\hypertarget{theoretical-background}{%
\section{Theoretical background}\label{theoretical-background}}

\label{section:theoretical}

In this section, the theoretical foundations of the NN2Poly algorithm
\citep{moralaNN2PolyPolynomialRepresentation2023a} will be presented
along with the needed notation. For further details, refer to the
original work.

\hypertarget{neural-networks-and-polynomials}{%
\subsection{Neural networks and
polynomials}\label{neural-networks-and-polynomials}}

The NN2Poly algorithm seeks to obtain an approximation of a given
trained densely connected feed-forward neural networks, or Multi-Layer
Perceptron (MLP), by means of one or several polynomials in the original
variables. Therefore, the used notation for both neural networks and
polynomials is introduced here. Note that during the rest of the paper,
the term \emph{neural networks} will be used in general to denote
feed-forward MLPs.

We consider neural networks with \(L-1\) hidden layers and \(h_{l}\)
neurons at each layer \(l\). The input variables to the network are
denoted by \(\Vec{x} = (x_1, \dots, x_p)\), with dimension \(p\), and
the output response is \(\Vec{y} = (y_1, \dots, y_c)\), with \(c=1\), if
there is a single output (usual regression setting), and \(c>1\) if
there is more than one output (usual classification setting with \(c\)
classes). At any given layer \(l\) and neuron \(j\), its output is
\(\prescript{(l)}{}{y}_j\), while its inputs are the outputs from the
previous layer, \(\prescript{(l-1)}{}{y}_i\) for
\(i \in 1, \dots, h_{l-1}\). By definition, the input to the first
hidden layer is the network's input, \(\prescript{(0)}{}{y}_i = x_i\)
for \(i \in 1, \dots, p\). The network's final output is denoted by
\(\prescript{(L)}{}{y}_j\) for \(i \in 1, \dots, c\).
\(\prescript{(l)}{}{W}\) denotes the weights matrix connecting layer
\(l-1\) to layer \(l\), where its element at row \(i\) and column \(j\)
is denoted by \(\prescript{(l)}{}{w}_{ij}\), and
\(\prescript{(l)}{}{g}\) is the activation function at that layer. Then,
the output from each neuron \(j = 1, \dots, h_l\) at layer \(l\) can be
written as follows: \begin{equation}
    \prescript{(l)}{}{y}_j=\prescript{(l)}{}{g}\left(\prescript{(l)}{}{u}_j\right)=\prescript{(l)}{}{g}\left(\sum_{i=0}^{h_{l-1}}\prescript{(l)}{}{w}_{ij}\prescript{(l-1)}{}{y}_i\right),
    \label{eq_neuron_computation}
\end{equation} where \(\prescript{(l)}{}{u}_j\) is the synaptic
potential, i.e., the value computed at the neuron before applying the
activation function. Note that the matrix \(\prescript{(l)}{}{W}\) has
dimensions \((h_{l-1} + 1)\times h_{l}\), including the bias term
\(\prescript{(l-1)}{}{y}_0 = 1\).

Polynomials will be denoted in general by \(P\), and will be of the
following form: \begin{equation}
P = \beta_{0} + \underbrace{\beta_{1} x_{1} + \dots + \beta_{p}x_{p}}_{1\mbox{-order interactions}} +\dots + \beta_{12} x_{1}x_{2} + \dots + \underbrace{\beta_{1\dots 1}x_1^{Q} + \dots + \beta_{p\dots p}x_p^{Q} }_{Q\mbox{-order interactions}},
\label{eq_poly_k}
\end{equation} where \(p\) is the number of variables \(x_i\) and \(Q\)
is the total order of the polynomial.

To simplify this notation, each monomial of order \(T\) can be
represented by a vector \(\Vec{t}=(t_1,t_2,\dots,t_p)\), where each
element \(t_i\) is an integer that represents the number of times that
variable \(i\) appears in the monomial, i.e., the multiplicity of that
variable, for all \(i=1,\dots,p\). Note that \(T=\sum_{i=1}^{p} t_i\).
As an example, the monomial containing the combination of variables
\(x_1^2 x_2 x_4\) in a polynomial with \(p=4\) will be:
\begin{equation*}
    B_{(2,1,0,1)} = \beta_{(2,1,0,1)} \cdot x_1^2 x_2 x_4,
\end{equation*} where \(B\) is used for the monomial, i.e., the
coefficient and the variables as a whole, and \(\beta\) is used for the
coefficient itself.

Using this representation, Equation \ref{eq_poly_k} can be written using
the multiplicities as: \begin{equation}
    P = \sum_{\Vec{t} \in \mathcal{T}(p,Q)} B_{\Vec{t}} = \sum_{\Vec{t} \in \mathcal{T}(p,Q)} \beta_{\Vec{t}} \cdot x_{1}^{t_1}\dots x_{p}^{t_p},
    \label{eq_poly_vec_t}
\end{equation} where \(\mathcal{T}(p,Q)\) is the set of all possible
vectors \(\Vec{t}\), for a given number of variables \(p\) and total
order \(Q\), and considering all interactions. Note that its cardinality
is \(|\mathcal{T}(p,Q)|\) the total number of terms in the polynomial,
which will be denoted by \(N_{p,Q}\). For convenience, the intercept
will be denoted as \(\beta_{(0)}\).

\hypertarget{nn2poly-method}{%
\subsection{NN2Poly method}\label{nn2poly-method}}

The goal is to find a polynomial representation for a neural network,
particularly to handle the non-linear activations of hidden neurons. For
MLPs with a single hidden layer, a solution was introduced in
\citep{moralaMathematicalFrameworkInform2021}, focusing on regression
problems. This method uses Taylor expansion on the activation functions
and applies combinatorial properties to determine the coefficients for
each variable combination in the final output. Since Taylor expansions
result in polynomials and the linear combinations of neurons can also be
represented as polynomials, this approach effectively creates a
polynomial that mirrors the neural network's behavior.

The extension of this idea to arbitrarily deep neural networks and both
classification and regression problems was provided by the NN2Poly
method in \citep{moralaNN2PolyPolynomialRepresentation2023a}. In this
work, the polynomials are built following a similar approach where each
neuron at each layer is represented as a polynomial. This method is
based on an iterative approach, which will yield two polynomials at each
neuron \(j\) and layer \(l\), one as the input to the non-linearity or
activation function (denoted by the prefix \({\mathrm{in}}\)), and a
second one as the output (denoted by the prefix \({\mathrm{out}}\)):
\begin{equation}
    \prescript{(l)}{\mathrm{in}}{P}_j =\sum_{\Vec{t} \in \mathcal{T}(p,Q)} \prescript{(l)}{\mathrm{in}}{B}_{j,\Vec{t}},\qquad\text{and}\qquad \prescript{(l)}{\mathrm{out}}{P}_j =\sum_{\Vec{t} \in \mathcal{T}(p,Q)} \prescript{(l)}{\mathrm{out}}{B}_{j,\Vec{t}},
\end{equation} following the notation from Equation \ref{eq_poly_vec_t}.
Note that the two polynomials at the same layer will be related by the
application of the activation function at such a layer,
\(\prescript{(l)}{}{g}()\), as follows: \begin{equation}
    \prescript{(l)}{\mathrm{out}}{P}_j\approx \prescript{(l)}{}{g}\left(\prescript{(l)}{\mathrm{in}}{P}_j\right).
\end{equation}

The key step is to be able to perform Taylor expansion of the activation
function, but considering another polynomial as input to that function.
The result used to perform this approximation is as follows:

\begin{lemma}
    Let $\prescript{}{\mathrm{in}}{P}$ be a polynomial of order $Q$ in $p$ variables and $g()$ a k-times differentiable function. Applying a Taylor expansion around $0$ and up to order $q$ yields a new polynomial $\prescript{}{\mathrm{out}}{P}$ in the same variables, of order $Q \times q$, whose coefficients are given by
    \begin{equation}
        \prescript{}{\mathrm{out}}{\beta}_{\vec{t}} = 
        \sum_{n=0}^{q}\dfrac{g^{(n)}(0)}{n!}
        \sum_{ \Vec{n} \in \pi(\Vec{t},Q,n)}
        \left(\begin{array}{c} n \\ \Vec{n} \end{array}\right)\prod_{k=1}^{N_{p,Q}} \prescript{}{\mathrm{in}}{\beta}_{k}^{n_{k}},
    \end{equation}
    for all $\Vec{t} \in \mathcal{T}(p,Q \times q)$, where $\pi(\Vec{t},Q,n)$ is the set of vectors $\Vec{n}=(n_1,\dots, n_{N_{p,Q}})$ that represents all the possible combinations of terms $\prescript{}{\mathrm{in}}{\beta}$ needed to obtain $\prescript{}{\mathrm{out}}{\beta}$, satisfying the following conditions:
\begin{itemize}
    \item \textbf{Condition 1}: $n_{1}+\dots+n_{M}=n$. 
    \item \textbf{Condition 2}: For all $k$, the order $T_k$ of the monomial $\prescript{}{\mathrm{in}}{B}_k$ associated to $\prescript{}{\mathrm{in}}{\beta}_k$  must satisfy that $T_k  \le Q$.
\end{itemize}
    \label{lemma_taylor_poly}
\end{lemma}

Note in the previous lemma that the polynomial coefficients inside the
product are represented as
\(\prescript{}{\mathrm{in}}{\beta}_{k}^{n_{k}}\) for all
\(k=1, \dots, N_{p,Q}\) instead of with the vector
\(\Vec{t}=(t_1,t_2,\dots,t_p)\) notation, but both of them are
equivalent and the \(k\) notation allows the product to be written as
\(\prod_{k=1}^{N_{p,Q}}\). Further details in
\citep{moralaNN2PolyPolynomialRepresentation2023a}.

The final algorithm implementation of the NN2Poly algorithm is given in
Algorithm \ref{alg_NN2Poly_practical}:

\begin{algorithm}[t]
\small
\caption{NN2Poly Algorithm}\label{alg_NN2Poly_practical}
\begin{algorithmic}[1]
\Require Weight matrices $\prescript{(l)}{}{W}$, activation functions $\prescript{(l)}{}{g}$,  Taylor truncation order at each layer $q_l$, maximum order $Q_{\mathrm{max}}$.
\State \textbf{Compute} all multiset partitions for every equivalent $M_0$ needed for a polynomial of order $Q_{\mathrm{max}}$ in $p$ variables.

\State Set $\prescript{(1)}{\mathrm{in}}{P}_j = \prescript{(1)}{}{u}_j$ for all $j=1, \dots, h_1$.
\For{$l=1, \cdots, L-1$}
    \For{$j=1, \cdots, h_l$} 
        \State \textbf{Compute} the coefficients, only up to order $Q_{l}^{*}$, from  $\prescript{(l)}{\mathrm{out}}{P}_j\approx \ g\left(\prescript{(l)}{\mathrm{in}}{P}_j\right)$ using Lemma \ref{lemma_taylor_poly}.
    \EndFor
    \For{$j=1, \cdots, h_{l+1}$} 
        \State \textbf{Compute} the coefficients, only up to order $Q_{l}^{*}$, from $\prescript{(l+1)}{\mathrm{in}}{P}_j = \sum_{i=0}^{h_{l}}\prescript{(l+1)}{}{w}_{ij}\prescript{(l)}{\mathrm{out}}{P}_i$
    \EndFor
\EndFor
\If{Non-linear output at last layer}
    \For{$j=1, \cdots, h_{L+1}$} 
        \State \textbf{Compute} the coefficients, only up to order $Q_{l}^{*}$, from  $\prescript{(L)}{\mathrm{out}}{P}_j\approx  g\left(\prescript{(L)}{\mathrm{in}}{P}_j\right)$ using Lemma \ref{lemma_taylor_poly}.
    \EndFor
\EndIf
\end{algorithmic}
\end{algorithm}

As shown in Algorithm \ref{alg_NN2Poly_practical}, a maximum order for
the obtained polynomial is set as \(Q_{\mathrm{max}}\), avoiding thus an
exponential increase in the number of interactions as the final
polynomial order would increase multiplicatively at each layer with the
chosen order for the Taylor expansion.

Furthermore, it is important to note that there need to be some
constraints on the original weights of the neural network, namely
keeping the weights vectors norm (including the bias) of the hidden
layers lower than a given value. This is implemented in \pkg{nn2poly},
as explained in Section \ref{section:constraints}.

\hypertarget{efficiency-improvements}{%
\subsection{Efficiency improvements}\label{efficiency-improvements}}

The proposed algorithm relies on being able to compute the set of
vectors \(\pi(\Vec{t},Q,n)\) defined in Lemma \ref{lemma_taylor_poly},
which is a non trivial combinatorial problem. For each vector
\(\Vec{t}\), this defines the set of all possible combinations of terms
in the previous polynomial \(\prescript{}{\mathrm{in}}{\beta}\) that,
when multiplied together, are needed to obtain the desired output
coefficient \(\prescript{}{\mathrm{out}}{\beta}\). This problem can be
converted in a combinatorial problem in terms of multisets. As explained
in \citet{moralaNN2PolyPolynomialRepresentation2023a}, each vector
\(\Vec{t} = (t_1,t_2,\dots,t_p)\) can be represented by a multiset \(M\)
given by \begin{equation}
    M = \left\{\underbrace{1,\dots,1}_{t_1}, \underbrace{2,\dots,2}_{t_2}, \dots, \underbrace{p,\dots,p}_{t_p}\right\},
    \label{eq_multiset_M_from_t}
\end{equation} where each variable \(i\) appears \(t_i\) times in the
multiset. Consider as an example the monomial \(x_1^2 x_2 x_4\). Then,
its associated coefficient using the vector \(\Vec{t}\) notation would
be \((2,1,0,1)\), and the associated multiset \(M\) would be
\(\{1,1,2,4\}\). With this multinomial notation, the problem of finding
all terms in the previous polynomial that, when multiplied, yield the
combination of variables denoted by \(M\) is reduced to finding all
possible partitions of said multiset \(M\). Finding all possible
partitions of a multiset can be done employing Algorithm 3 from
\citet{moralaNN2PolyPolynomialRepresentation2023a}, which is based in a
proposal from \citet{knuthArtComputerProgramming2005}. This method can
be employed on all multisets, but as the number of variables and order
of the polynomial grows, the number of polynomial coefficients rapidly
increases with a multiset associated to each of them. To mitigate the
computational burden of this situation, it can be noticed that there are
equivalencies in the monomials. For instance, computing the partitions
of multiset \(\{1,1,2,4\}\) is equivalent to computing them for multiset
\(\{2,2,3,5\}\) or \(\{1,3,5,5\}\), as long as the number of distinct
elements and their multiplicities are the same (see Section IV.C of
\citet{moralaNN2PolyPolynomialRepresentation2023a} for further details).

The manipulation of multinomial partitions and their equivalent
representations are implemented in \proglang{C++} to further reduce
computational times.

\hypertarget{package-structure}{%
\section{Package structure}\label{package-structure}}

\label{section:pkg_struc}

The package is built around one main function, \code{nn2poly()}, where
the algorithm of NN2Poly is implemented. This function receives a neural
network, and returns a polynomial represented as a new \code{S3} object
named \pkg{nn2poly}, also defined inside the package. Then,
\code{plot()} and \code{predict()} methods are implemented to be used on
the \pkg{nn2poly} class. Additionally, two functions are defined to
properly build and train neural networks with the desired constraints
and structure to be used with \code{nn2poly()}, namely
\code{add_constraints()} and \code{luz_model_sequential()}.

\hypertarget{polynomial-structure}{%
\subsection{Polynomial structure}\label{polynomial-structure}}

\label{section:polynomial_structure}

Given that the main objective with NN2Poly is to obtain a polynomial
model equivalent to the original neural network, we will introduce here
the structure used in the \pkg{nn2poly} package to define polynomials.
This structure will be used in the main function \code{nn2poly()}
output, but we first introduce it here as it can also be useful when
generating synthetic data for the given examples.

A single polynomial will be represented as a list with two elements,
named \code{labels} and \code{values}:

\begin{itemize}
\item
  Item \code{labels} is a list of integer vectors. Each of those vectors
  represents a monomial in the polynomial, where each integer in the
  vector represents each time one of the original variables appears in
  that term. As an example, vector \code{c(1,1,2)} represents the term
  \(x_1 \cdot x_1 \cdot x_2 = x_1^2x_2\). Note that the variables are
  numbered from \(1\) to \(p\), with the intercept being represented by
  \(0\).
\item
  Item \code{values} contains a column vector with the coefficients of
  the polynomial, where each element in the vector is the coefficient
  associated with the monomial or label at the same position in the
  \code{labels} item.
\end{itemize}

\clearpage

Let's see an example with the polynomial: \[
Y = 4 + 2X_1-X_2X_3+3 X_4X_5
\] which is built as follows:

\begin{CodeChunk}
\begin{CodeInput}
R> polynomial_simple <- list()
R> polynomial_simple$labels <- list(c(0), c(1), c(2, 3), c(4, 5))
R> polynomial_simple$values <- c(4, 2, -1, 3)
\end{CodeInput}
\end{CodeChunk}

If instead of a single vector we wanted to represent multiple vectors
with the same labels or monomials, item \code{values} can be a matrix
instead of a vector, where each column contains the coefficients for
each vector. For example, the following two polynomials: \[
Y = 4 + 2X_1-X_2X_3+3 X_4X_5
\] \[
Y = 1 - 3X_1+2X_2X_3+ X_4X_5
\]

\begin{CodeChunk}
\begin{CodeInput}
R> polynomial_multi <- list()
R> polynomial_multi$labels <- list(c(0), c(1), c(2, 3), c(4, 5))
R> polynomial_multi$values <- cbind(c(4, 2, -1, 3), c(1, -3, 2, 1))
\end{CodeInput}
\end{CodeChunk}

This representation of multiple polynomials with the same monomials but
different coefficients will be employed in the \pkg{nn2poly} output when
representing a neural network with multiple output units, as in a
classification problem, as in Section \ref{section:classification}.

\hypertarget{neural-network-input}{%
\subsection{Neural network input}\label{neural-network-input}}

The package currently supports 3 possible ways of using a neural network
as input: the default input object as a list of weight matrices, a
\pkg{tensorflow}-based \pkg{keras} sequential object or a
\pkg{torch}-based \pkg{luz} sequential object:

\hypertarget{generic-list-input}{%
\subsubsection{Generic list input}\label{generic-list-input}}

The \code{nn2poly()} function allows users to input any pre-trained
neural network from any framework. The input is a list of matrices, each
representing the weight matrices of the network layers in sequence.
These matrices must include bias vectors as the first row, with each
subsequent row representing the weights of input neurons at that layer.
For a layer \(l+1\), the weight matrix dimensions are
\((1+h_{l})\times h_{l+1}\) , where \(h_l\) is the number of neurons in
layer \(l\). The activation function for each layer is specified by the
name of the corresponding matrix in the list. Supported activation
functions include hyperbolic tangent, sigmoid, and softplus. The package
can implement other differentiable activation functions, though it
cannot handle the non-differentiable ReLU due to theoretical
constraints.

As an illustrative example of a list input to \code{nn2poly()}, suppose
a neural network with an input dimension of 5, two hidden layers
containing 2 and 3 neurons respectively, using hyperbolic tangent and
sigmoid activation functions, and an output layer with 1 neuron and a
linear activation function. Then, the list that represents the network
in order to use it as input of \code{nn2poly()} will have the following
structure (note that all layers have bias weights).

\begin{CodeChunk}
\begin{CodeInput}
R> # We will fill the weights as 0s for the bias and 1s for the kernel weights
R> # just to show an example of the required dimensions.
R> dim_input <- 5
R> dim_layer_1 <- 2
R> fun_layer_1 <- "tanh"
R> bias_weights_layer_1 <- matrix(0, nrow = 1, ncol = dim_layer_1)
R> kernel_weights_layer_1 <- matrix(1, nrow = dim_input, ncol = dim_layer_1)
R> 
R> dim_layer_2 <- 3
R> fun_layer_2 <- "sigmoid"
R> bias_weights_layer_2 <- matrix(0, nrow = 1, ncol = dim_layer_2)
R> kernel_weights_layer_2 <- matrix(1, nrow = dim_layer_1, ncol = dim_layer_2)
R> 
R> dim_output <- 1
R> fun_output <- "linear"
R> bias_weights_layer_output <- matrix(0, nrow = 1, ncol = dim_output)
R> kernel_weights_layer_output <- matrix(1, nrow = dim_layer_2, ncol = dim_output)
R> 
R> # Final input list:
R> nn_list <- list(
+   rbind(bias_weights_layer_1, kernel_weights_layer_1),
+   rbind(bias_weights_layer_2, kernel_weights_layer_2),
+   rbind(bias_weights_layer_output, kernel_weights_layer_output))
R> names(nn_list) <- c(fun_layer_1, fun_layer_2, fun_output)
R> nn_list
\end{CodeInput}
\begin{CodeOutput}
$tanh
     [,1] [,2]
[1,]    0    0
[2,]    1    1
[3,]    1    1
[4,]    1    1
[5,]    1    1
[6,]    1    1

$sigmoid
     [,1] [,2] [,3]
[1,]    0    0    0
[2,]    1    1    1
[3,]    1    1    1

$linear
     [,1]
[1,]    0
[2,]    1
[3,]    1
[4,]    1
\end{CodeOutput}
\end{CodeChunk}

Building this object with the appropriate format is left to the user
when wanting to apply \code{nn2poly()} to a neural network from a non
supported framework. However, the supported frameworks can be directly
used with \code{nn2poly()} as the adequate input format is built
internally for each method. Therefore, providing support for new deep
learning formats can be rapidly implemented by extending the internal
functions to get weights and activation functions from models of that
new framework's class. In particular, currently supported input formats
are sequential models in \pkg{keras} and \pkg{luz}. For each framework
we provide an example of a sequential model.

\subsubsection[keras sequential input]{\pkg{keras} sequential input}
\label{section:keras_seq_input}

\pkg{nn2poly} provides support for \pkg{tensorflow}/\pkg{keras} models
composed of a linear stack of layers, which can be built using the
\code{keras::keras_model_sequential()} function. These models can be
directly fed to \code{nn2poly()}, which internally extracts and reshapes
its needed parameters (weight matrices and activation function names) to
comply with the list format explained before.

As an illustrative example we will replicate here how to create a
\pkg{keras} network with the structure shown in the list input example:

\begin{CodeChunk}
\begin{CodeInput}
R> # keras required (which also requires tensorflow and Python installed)
R> library("keras")
R> tensorflow::set_random_seed(42)
R> 
R> nn_toy_keras <- keras_model_sequential() %>%
+   layer_dense(units = 2, activation = "tanh", input_shape = 5) %>%
+   layer_dense(units = 3, activation = "sigmoid") %>%
+   layer_dense(units = 1, activation = "linear")
R> nn_toy_keras
\end{CodeInput}
\begin{CodeOutput}
Model: "sequential"
________________________________________________________________________________
 Layer (type)                       Output Shape                    Param #     
================================================================================
 dense_2 (Dense)                    (None, 2)                       12          
 dense_1 (Dense)                    (None, 3)                       9           
 dense (Dense)                      (None, 1)                       4           
================================================================================
Total params: 25 (100.00 Byte)
Trainable params: 25 (100.00 Byte)
Non-trainable params: 0 (0.00 Byte)
________________________________________________________________________________
\end{CodeOutput}
\end{CodeChunk}

\subsubsection[luz sequential input]{\pkg{luz} sequential input}
\label{section:luz_seq_input}

The \pkg{nn2poly} package provides support for models in sequential
form, i.e., composed of a linear stack of layers. In this case, we
provide the helper function \code{luz_model_sequential()} to facilitate
the creation of such models. Then, they can be directly fed to
\code{nn2poly()}, which internally extracts and reshapes its needed
parameters (weight matrices and activation function names) to comply
with the list format explained before.

As an illustrative example, we will replicate the structure shown in the
list input example: a neural network with an input dimension of 5, two
hidden layers containing 2 and 3 neurons respectively, using hyperbolic
tangent and sigmoid activation functions, and an output layer with 1
neuron and a linear activation function.

\begin{CodeChunk}
\begin{CodeInput}
R> # luz and torch required
R> library("luz")
R> library("torch")
R> 
R> # Create a luz sequential model with the helper function
R> nn_toy_luz <- nn2poly::luz_model_sequential(
+   nn_linear(5, 2),
+   nn_tanh(),
+   nn_linear(2, 3),
+   nn_sigmoid(),
+   nn_linear(3, 1)
+ )
R> nn_toy_luz
\end{CodeInput}
\begin{CodeOutput}
<nn_sequential> object generator
  Inherits from: <inherit>
  Public:
    .classes: nn_sequential nn_module
    args: list
    initialize: function () 
    forward: function (input) 
    clone: function (deep = FALSE, ..., replace_values = TRUE) 
  Private:
    .__clone_r6__: function (deep = FALSE) 
  Parent env: <environment: 0x564385a680b8>
  Locked objects: FALSE
  Locked class: FALSE
  Portable: TRUE
\end{CodeOutput}
\end{CodeChunk}

\hypertarget{neural-network-training-with-constraints}{%
\subsection{Neural network training with
constraints}\label{neural-network-training-with-constraints}}

\label{section:constraints}

Before getting into how the \code{nn2poly()} function is implemented, we
introduce a feature that allows the use of weight constraints during the
neural network training. The NN2Poly theoretical method depends on the
effectiveness of the Taylor expansion performed at each neuron.
Therefore, if the neural networks is trained without any constraints on
the weights, the obtained polynomial may not be accurate or even greatly
diverge on its predictions due to an asymptotic behavior on the Taylor
expansions. To avoid this, weight constraints can be used as explained
in Section \ref{section:theoretical}, where the norm of weight vectors
at hidden layers can be limited to have absolute value lower than one.
This condition has to be imposed during training, and therefore we
provide an implementation to add such a restriction in the supported
deep learning frameworks.

To do so, the package provides function \code{add_constraints()} that
can be used on any sequential neural network from the supported
frameworks explained previously (Tensorflow+Keras or Torch+Luz). The
approach used to introduce the constraints consists of using callbacks
that are employed during training when using fit on the model.
Specifically, the callbacks are applied at the end of each train batch,
on each weight vector (columns in the weight matrices in our standard
format, thus representing the weights incident on a neuron, with the
bias as the first element and the rest of the weights sequentially in
the rest of the vector). For each of those vectors \(w\), a new weight
vector is computed as follows: \[
\vec{w}_{new} = \vec{w} \times \dfrac{c\left(||\vec{w}||\right)}{||\vec{w}||+\epsilon}
\] where function \(c()\) is defined as as: \[
c(x) = \begin{cases}
    x & \text{if } x \in [0,1] \\
    1 & \text{if } x > 1.
\end{cases}
\]

This computation leaves the weight vector unchanged if its norm is less
than 1, and scales it to a new vector with a norm of 1 if its previous
norm was greater than 1. The implemented norms are the \emph{l1-norm}
and the \emph{l2-norm}: \[
||\vec{w}||_1 = \sum_{i=0}^p \left\lvert {w}_{i} \right\rvert \quad\text{and}\quad ||\vec{w}||_2 =  \sqrt{\sum_{i=0}^p {w}_{i}^2}.
\]

Using function \code{add_constraints(model, type)} adds class
\code{"nn2poly"} to the given model, and a new model attribute named
\code{"constraint"} which stores the type of norm to be used during
training (either \code{"l1_norm"} or \code{"l2_norm"}). This allows the
use of method \code{fit.nn2poly()}, which builds the callback before
calling the appropriate \code{fit()} method for each given model, where
the training will be performed as usual within the models framework but
with the desired constraint included as a callback in said framework.

Note that, if the batch size is too small, the computational time of the
callbacks may be of similar magnitude as the batch training time, which
may produce some warnings depending on the used framework.

\hypertarget{data-generation}{%
\subsubsection{Data generation}\label{data-generation}}

In order to show an example on how this constraints are used during
training with each supported framework, we will first generate some
synthetic data. This data will be of polynomial nature for the sake of
being able to compare the results obtained later with \code{nn2poly()}.
Following the polynomial notation defined in Section
\ref{section:polynomial_structure}, the example polynomial will be: \[
Y = 2 - 2X_1 + 5X_2X_3 + 3X_4
\]

\begin{CodeChunk}
\begin{CodeInput}
R> # Polynomial with interactions
R> poly <- list()
R> poly$labels <- list(c(0), c(1), c(2, 3), c(4))
R> poly$values <- c(2, -2, 5, 3)
\end{CodeInput}
\end{CodeChunk}

Note how only four variables are considered in the polynomial, however,
data will be generated with dimension \(p=5\). Then, as variable \(X_5\)
does not appear in the polynomial, the neural network should not capture
any effect and therefore it should neither appear in the explanations.

To demonstrate data generation from this polynomial, we will use here an
internal function of \pkg{nn2poly}, namely \code{nn2poly:::eval_poly()},
which takes a polynomial in the previously defined structure and
evaluates it on the given data. Note that, as this is an internal
function, it may not be supported with the same syntax in future
versions.

\begin{CodeChunk}
\begin{CodeInput}
R> # Define number of variables and sample size
R> set.seed(42)
R> p <- 5
R> n <- 500
R> 
R> # Predictor variables X and response variable Y + small error term
R> x <- matrix(rnorm(n * p, 0, 1), n, p)
R> y <- nn2poly:::eval_poly(poly = poly, newdata = x) + rnorm(n, 0, 0.05)
R> data_reg <- as.data.frame(cbind(x, y))
R> head(data_reg)
\end{CodeInput}
\begin{CodeOutput}
          V1           V2         V3         V4         V5         y
1  1.3709584  1.029140719  2.3250585 -0.6013830  0.2505781  9.448863
2 -0.5646982  0.914774868  0.5241222 -0.1358161 -0.2779240  5.118990
3  0.3631284 -0.002456267  0.9707334 -0.9872728 -1.7247357 -1.704560
4  0.6328626  0.136009552  0.3769734  0.8319250 -2.0067049  3.506408
5  0.4042683 -0.720153545 -0.9959334 -0.7950595 -1.2918083  2.421855
6 -0.1061245 -0.198124330 -0.5974829  0.3404646  0.3658382  3.824678
\end{CodeOutput}
\end{CodeChunk}

In order to properly use this data when training a neural network, we
will first scale the data to the \([-1,1]\) interval and split in train
and test datasets:

\begin{CodeChunk}
\begin{CodeInput}
R> # Data scaling to [-1,1]
R> maxs <- apply(data_reg, 2, max)
R> mins <- apply(data_reg, 2, min)
R> data_reg <- as.data.frame(scale(data_reg, 
+                                 center = mins + (maxs - mins) / 2, 
+                                 scale = (maxs - mins) / 2))
R> 
R> # Divide in train (0.75) and test (0.25)
R> set.seed(42)
R> i <- sample(1:nrow(data_reg), round(0.75 * nrow(data_reg)))
R> train_reg_x <- as.matrix(data_reg[ i, ][, -(p+1)])
R> train_reg_y <- as.matrix(data_reg[ i, ][,  (p+1)])
R> test_reg_x  <- as.matrix(data_reg[-i, ][, -(p+1)])
R> test_reg_y  <- as.matrix(data_reg[-i, ][,  (p+1)])
\end{CodeInput}
\end{CodeChunk}

\clearpage

\hypertarget{constraints-and-training}{%
\subsubsection{Constraints and
training}\label{constraints-and-training}}

Here we will build a \pkg{keras} model and then add the required
constraints to it using the method provided by \pkg{nn2poly}.

\begin{CodeChunk}
\begin{CodeInput}
R> library("nn2poly")
R> tensorflow::set_random_seed(42)
R> 
R> nn_reg <- keras_model_sequential() %>%
+   layer_dense(units = 50, activation = "tanh", input_shape = 5) %>%
+   layer_dense(units = 100, activation = "tanh") %>%
+   layer_dense(units = 50, activation = "tanh") %>%
+   layer_dense(units = 1, activation = "linear")
R> 
R> # Add the constraints
R> nn_reg_const <- add_constraints(nn_reg, type = "l1_norm")
R> 
R> # We can see how it preserves the original structure:
R> nn_reg_const
\end{CodeInput}
\begin{CodeOutput}
Model: "sequential_1"
________________________________________________________________________________
 Layer (type)                       Output Shape                    Param #     
================================================================================
 dense_6 (Dense)                    (None, 50)                      300         
 dense_5 (Dense)                    (None, 100)                     5100        
 dense_4 (Dense)                    (None, 50)                      5050        
 dense_3 (Dense)                    (None, 1)                       51          
================================================================================
Total params: 10501 (41.02 KB)
Trainable params: 10501 (41.02 KB)
Non-trainable params: 0 (0.00 Byte)
________________________________________________________________________________
\end{CodeOutput}
\begin{CodeInput}
R> # See how "nn2poly" class is added
R> setdiff(class(nn_reg_const), class(nn_reg))
\end{CodeInput}
\begin{CodeOutput}
[1] "nn2poly"
\end{CodeOutput}
\begin{CodeInput}
R> # Show the kind of constraint to be used, stored as an attribute
R> # When type is not specified, defaults to "l1_norm"
R> attributes(nn_reg_const)$constraint
\end{CodeInput}
\begin{CodeOutput}
[1] "l1_norm"
\end{CodeOutput}
\end{CodeChunk}

Due to the \proglang{Python} internal nature of the \pkg{keras} and
\pkg{tensorflow} packages, implementing the constraints within
\proglang{R} incurs in a slowdown of the training procedure, because of
the back and forth at the end of each batch between \proglang{Python}
and \proglang{R} to compute the constraint. Therefore, for the
\pkg{tensorflow}+\pkg{keras} framework, the constraints have been
implemented inside \proglang{Python} as classes \code{KerasCallback} and
\code{KerasConstraint}. Now we can train the model as usual within
\pkg{keras}:

\begin{CodeChunk}
\begin{CodeInput}
R> compile(nn_reg_const, optimizer = optimizer_adam(),
+         loss = "mse", metrics = "mse")
R> history <- fit(nn_reg_const, train_reg_x, train_reg_y, verbose = FALSE,
+                epochs = 900, batch_size = 50, validation_split = 0.2)
R> plot(history) + ggplot2::facet_grid(~metric)
\end{CodeInput}
\begin{figure}

{\centering \includegraphics[width=1\linewidth]{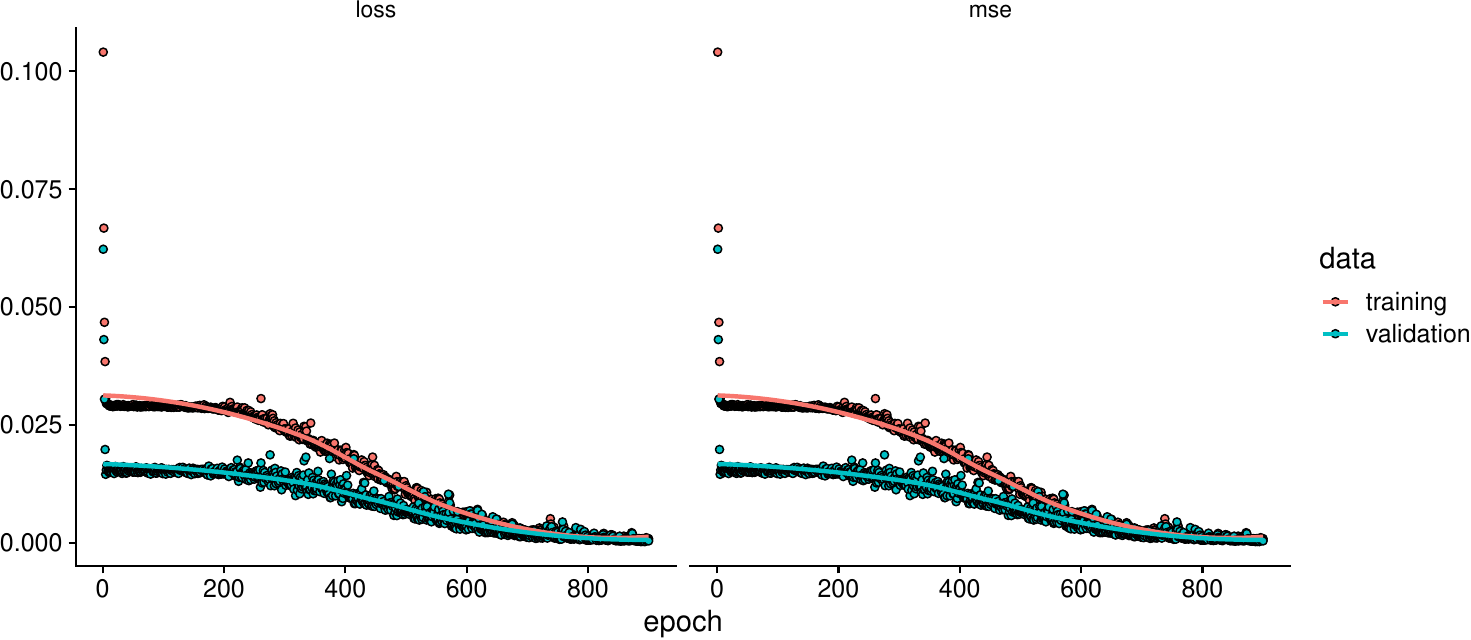} 

}

\caption[Training history of the neural network with constraints for the regression model]{Training history of the neural network with constraints for the regression model.}\label{fig:training-keras-history}
\end{figure}
\end{CodeChunk}

Figure \ref{fig:training-keras-history} shows how the imposed
constraints limit the model learning at first, with around the 200 first
epochs showing a flat trend. However, after that moment, the loss
function starts decreasing and the learning process converges to an
optimal solution. This showcases the trade-off between learning speed
and the interpretability achieved by using this constraints with our
method.

As expected, the 3 (constrained) hidden layers have weights with an
\emph{l1-norm} lower or equal than 1, while the (unconstrained) output
layer does not:

\begin{CodeChunk}
\begin{CodeInput}
R> # Extract the parameters in the needed manner (internal nn2poly function used)
R> params <- nn2poly:::get_parameters.keras.engine.training.Model(nn_reg_const)
R> 
R> # Compute the norm of each vector in each layer
R> # These are the columns of our weight matrices.
R> weights_norms <- lapply(params$weights_list, apply, 2, pracma::Norm, 1)
R> 
R> # Check which layers have weight vectors with norm >=1
R> sapply(weights_norms, function(x) all(x <= 1))
\end{CodeInput}
\begin{CodeOutput}
[1]  TRUE  TRUE  TRUE FALSE
\end{CodeOutput}
\end{CodeChunk}

The process is completely equivalent for the \pkg{luz}+\pkg{torch}
framework, with the addition of the constraints between the
\pkg{setup()} and \pkg{fit} phases. The only difference is that a
specific data loader is required as input (see the specific
documentation for more details).

\hypertarget{extracting-the-polynomial-representation}{%
\subsection{Extracting the polynomial
representation}\label{extracting-the-polynomial-representation}}

\label{section:main_nn2poly}

Once we have a neural network trained with the desired constraints and
in an appropriate input form, obtaining the polynomial representation is
done through the \code{nn2poly()} function. Different neural network
frameworks have been explained before this point, but all of them can be
used in \code{nn2poly()}'s argument. Hence, from this point onward, we
will use the \pkg{keras} neural network to demonstrate the rest of the
package's functionality.

Using the \code{nn2poly()} function, a polynomial of order determined by
parameter \code{max_order} is obtained, in the same shape as explained
in Section \ref{section:polynomial_structure}:

\begin{CodeChunk}
\begin{CodeInput}
R> poly_nn <- nn2poly(nn_reg_const, max_order = 3)
R> names(poly_nn)
\end{CodeInput}
\begin{CodeOutput}
[1] "labels" "values"
\end{CodeOutput}
\begin{CodeInput}
R> rle(sapply(poly_nn$labels, length))
\end{CodeInput}
\begin{CodeOutput}
Run Length Encoding
  lengths: int [1:3] 6 15 35
  values : int [1:3] 1 2 3
\end{CodeOutput}
\end{CodeChunk}

Note that we have chosen a maximum order of 3 for the network trained on
an order 2 polynomial. This approach will be useful to see if those
higher-order terms converge to 0. Obviously, the needed order is not
known previously in a real dataset application, but the order can always
be increased if the polynomial does not predict similarly to the
original network. As expected, for 5 variables, we obtain 6 terms of
order 1 (intercept + 5 variables), 15 terms of order 2, and 35 terms of
order 3 (combinations with repetition).

An additional feature in the \code{nn2poly()} function is that it can
also provide all the internal polynomial representations obtained
through each step from Algorithm \ref{alg_NN2Poly_practical} when
setting \code{keep_layers = TRUE}, where an input polynomial and an
output polynomial is obtained at each layer, with the same encoding as
before:

\begin{CodeChunk}
\begin{CodeInput}
R> poly_internal <- nn2poly(nn_reg_const, max_order = 3, keep_layers=TRUE)
R> str(poly_internal$layer_2, 2)
\end{CodeInput}
\begin{CodeOutput}
List of 2
 $ input :List of 2
  ..$ labels:List of 56
  ..$ values: num [1:56, 1:100] -0.00461 0.00493 0.0025 -0.00137 -0.0069 ...
 $ output:List of 2
  ..$ labels:List of 56
  ..$ values: num [1:56, 1:100] -0.00461 0.00493 0.0025 -0.00137 -0.0069 ...
\end{CodeOutput}
\end{CodeChunk}

By the algorithm construction, the labels of the polynomial terms remain
the same at each layer, with only their values changing. Therefore, the
multiple polynomial representation introduced in Section
\ref{section:polynomial_structure} is used. In this representation, each
column of the \code{$values} matrix represents a different
polynomial---in this case, three, corresponding to the neurons present
at layer 2 in \code{nn_keras}. These internal polynomials have the
potential use of explaining specific parts of the neural network instead
of only the output.

\hypertarget{making-predictions}{%
\subsection{Making predictions}\label{making-predictions}}

Obtaining predictions with the final polynomials is also possible by
using the \code{predict()} method. Here we obtain predictions from the
trained neural network as well as from our polynomial representation:

\begin{CodeChunk}
\begin{CodeInput}
R> prediction_nn <- predict(nn_reg_const, x = test_reg_x, verbose = FALSE)
R> prediction_poly_nn  <- predict(poly_nn, newdata = test_reg_x)
\end{CodeInput}
\end{CodeChunk}

The polynomial predictions can be compared with the original neural
network predictions to asses the equivalency in terms of predictions of
the model. As it can be seen in Figure \ref{fig:nn2poly-compare}, the
polynomial predictions obtained with \pkg{nn2poly} are almost equal to
the ones obtained by the neural network. It should be noted here that we
do not compare the polynomial predictions with the original response
value, as the polynomial predictions should mimic the neural network
independently of how well it has learnt from the original data.

\begin{CodeChunk}
\begin{figure}

{\centering \includegraphics[width=0.495\linewidth]{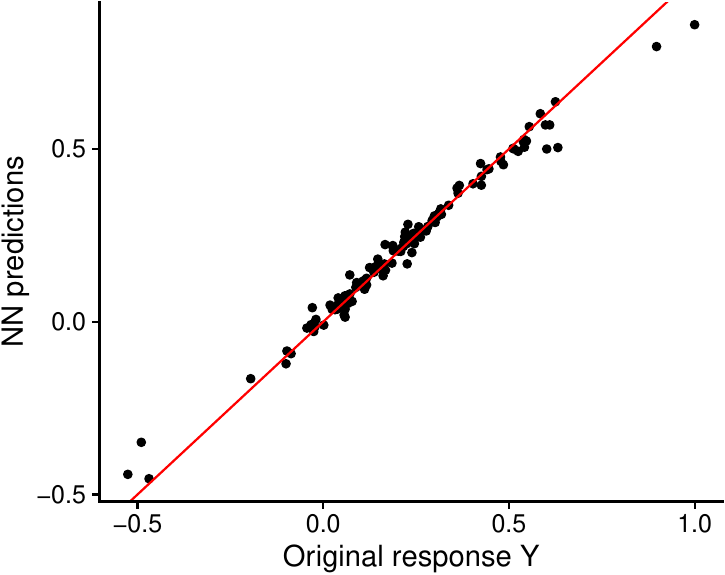} \includegraphics[width=0.495\linewidth]{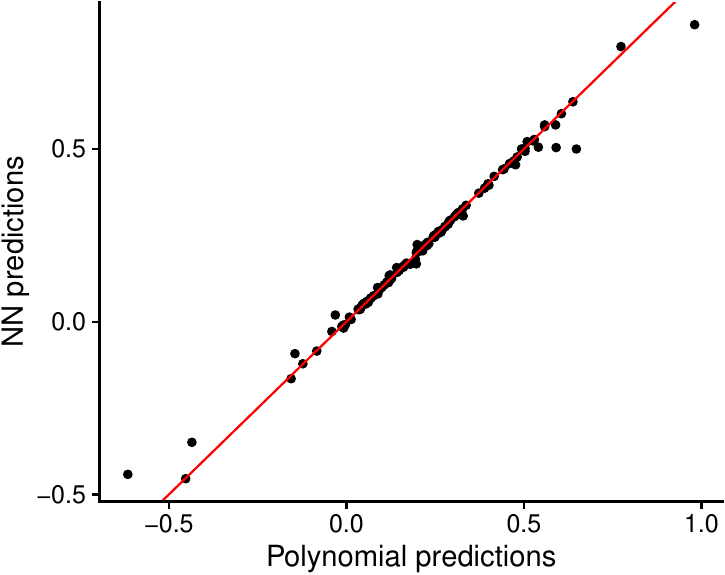} 

}

\caption[Predictions from the neural network model compared with
(left) the original response variable and
(right) the preditions from the polynomial representation]{Predictions from the neural network model compared with
(left) the original response variable and
(right) the preditions from the polynomial representation.}\label{fig:nn2poly-compare}
\end{figure}
\end{CodeChunk}

\hypertarget{interpretation-via-polynomial-coefficients}{%
\subsection{Interpretation via polynomial
coefficients}\label{interpretation-via-polynomial-coefficients}}

Outputs from \code{nn2poly()} can be easily visualized by using the
\code{plot()} method:

\begin{CodeChunk}
\begin{CodeInput}
R> plot(poly_nn, n = 10)
\end{CodeInput}
\begin{figure}

{\centering \includegraphics[width=0.5\linewidth]{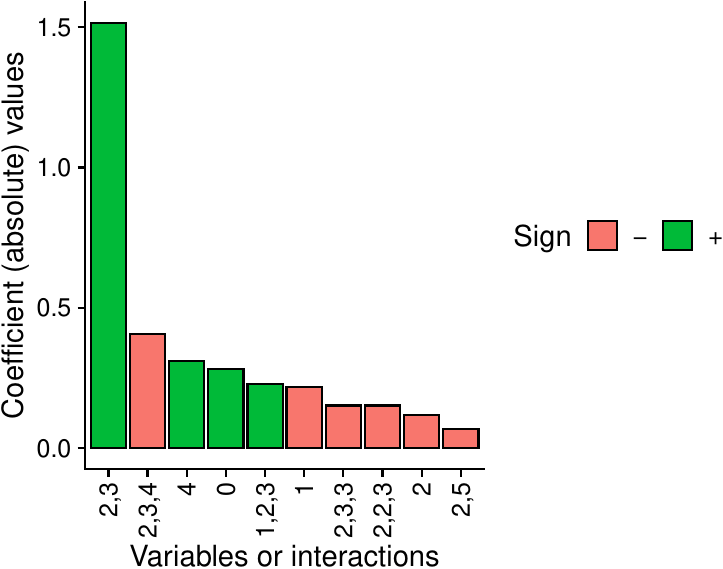} 

}

\caption[Top 10 most important coefficients from the polynomial representation]{Top 10 most important coefficients from the polynomial representation.}\label{fig:nn2poly-plot}
\end{figure}
\end{CodeChunk}

The original polynomial had an interaction between variables 2 and 3
(\(\beta_{2,3}= 5\)), which gets a strong positive coefficient in Figure
\ref{fig:nn2poly-plot}, while the single variables appearing in the
original polynomial (\(\beta_1 =-2\) and \(\beta_4 =3\)) have smaller
coefficients with their correct sign assigned. Note that the coefficient
values are not the same as in the original polynomial, as the data has
been scaled to the \([-1,1]\) interval. We can also observe that some
third order interactions appear with some importance, such as
\(\beta_{2,3,4}\), \(\beta_{1,2,3}\) or \(\beta_{2,3,3}\), all of them
including variables 2 and 3, and therefore capturing some of the
interaction between them. In this case, the increased complexity of the
data due to those interactions makes the neural network model to learn
some spurious relationships, which are reflected the final obtained
polynomial. Furthermore, note that higher-order coefficients would need
larger values to have a meaningful impact as, again, the data has been
scaled to the \([-1,1]\) interval.

\hypertarget{classification-example}{%
\subsection{Classification example}\label{classification-example}}

\label{section:classification}

The previous step-by-step examples were focused on solving a regression
task on a given polynomial, but \pkg{nn2poly} is not limited to such
case. Classification problems with neural networks mainly consist of a
neural network with several outputs, where each output neuron provides a
value between 0 and 1 representing the probability of the given
observation belonging to each possible class. Then, the class with the
highest value is assigned as the final prediction. However, those
neurons compute a linear output before applying the needed activation
function to transform it into a probability. Precisely, nn2poly can be
used to obtain a polynomial representing that linear output.

To exemplify this, here we will use a neural network to classify three
different species (Adelie, Chinstrap and Gentoo) of penguins based some
lengths of their body parts and their weight, available in
\pkg{palmerpenguins} \citep{R-palmerpenguins}. Then we will use
\pkg{nn2poly} to find a polynomial for each class.

First, we need to prepare the data by scaling the predictor variables to
the \([-1,1]\) interval.

\begin{CodeChunk}
\begin{CodeInput}
R> penguins <- na.omit(palmerpenguins::penguins)
R> 
R> # Scale the data in the [-1,1] interval and separate train and test
R> # Only the predictor variables are scaled, not the response as those will be
R> # the different classes.
R> penguins_x <- penguins[, c("bill_length_mm", "bill_depth_mm",
+                            "flipper_length_mm", "body_mass_g")]
R> maxs <- apply(penguins_x, 2, max)
R> mins <- apply(penguins_x, 2, min)
R> data_cls <- as.data.frame(scale(penguins_x,
+                                 center = mins + (maxs - mins) / 2,
+                                 scale = (maxs - mins) / 2))
R> 
R> # Species need to be transformed into numbers from 0 to n_cls-1, as 
R> # the keras/tensorflow is trained in Python and vectors start at 0.
R> penguins_y <- as.numeric(factor(penguins$species)) - 1
R> 
R> p <- dim(penguins_x)[2]
R> n_cls <- length(unique(penguins_y))
R> 
R> # Joint X and Y
R> data_cls <- cbind(data_cls, penguins_y)
R> 
R> # Divide in train (0.75) and test (0.25)
R> i <- sample(1:nrow(data_cls), round(0.75 * nrow(data_cls)))
R> train_cls_x <- as.matrix(data_cls[ i, ][, -(p+1)])
R> train_cls_y <- as.matrix(data_cls[ i, ][,  (p+1)])
R> test_cls_x  <- as.matrix(data_cls[-i, ][, -(p+1)])
R> test_cls_y  <- as.matrix(data_cls[-i, ][,  (p+1)])
\end{CodeInput}
\end{CodeChunk}

Then, we can build a \pkg{keras} neural network with 3 output neurons in
this case, corresponding to the three different species of penguins
present in the dataset. Note how the networks output is still linear,
but we define a \emph{sparse categorical crossentropy} loss function to
work in a classification problem. Predictions will then include a
\emph{softmax} layer, but this is not needed during training.

\begin{CodeChunk}
\begin{CodeInput}
R> tensorflow::set_random_seed(42)
R> 
R> nn_cls <- keras_model_sequential() %>%
+   layer_dense(units = 100, activation = "tanh", input_shape = p) %>%
+   layer_dense(units = 100, activation = "tanh") %>%
+   layer_dense(units = n_cls, activation = "linear") %>%
+   add_constraints(type = "l1_norm") %>%
+   compile(loss = loss_sparse_categorical_crossentropy(from_logits = TRUE),
+           optimizer = optimizer_adam(),
+           metrics = "accuracy")
R> 
R> history <- fit(nn_cls, train_cls_x, train_cls_y, verbose = FALSE,
+                epochs = 100, validation_split = 0.3)
R> plot(history) + ggplot2::facet_grid(~metric)
\end{CodeInput}
\begin{figure}

{\centering \includegraphics[width=1\linewidth]{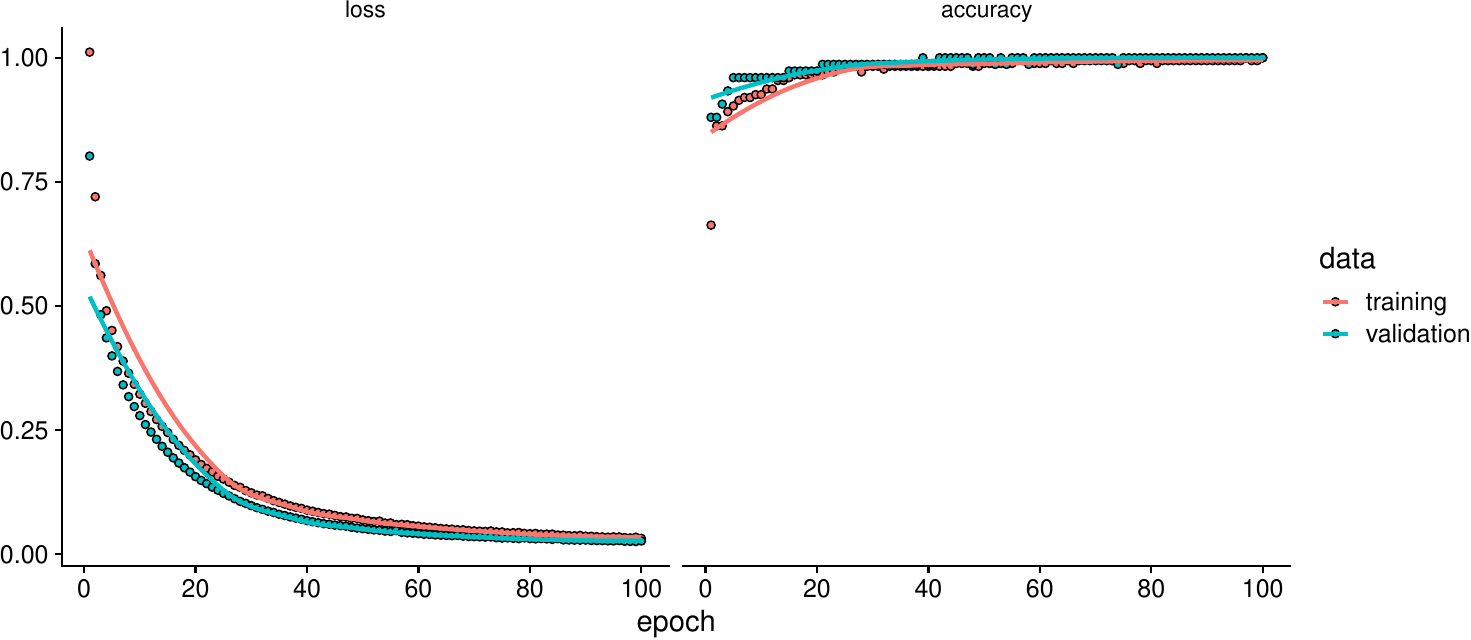} 

}

\caption[Training history of the neural network with constraints for the classification model]{Training history of the neural network with constraints for the classification model.}\label{fig:classification-keras}
\end{figure}
\end{CodeChunk}

Once the network has been trained (see Figure
\ref{fig:classification-keras} for the training history),
\code{nn2poly()} can be used to transform the model into polynomials for
each of the output nodes, i.e.~for each of the penguin species.

\begin{CodeChunk}
\begin{CodeInput}
R> # Polynomial for nn
R> poly_nn_cls <- nn2poly(nn_cls, max_order = 3)
R> 
R> # Check that the number of columns in the polynomial values matrix is
R> # equal to the number of classes to predict
R> dim(poly_nn_cls$values)[2]
\end{CodeInput}
\begin{CodeOutput}
[1] 3
\end{CodeOutput}
\end{CodeChunk}

However, the polynomials represent the model with linear output
(\code{nn}), but the actual classifications into the desired penguins
species are provided by some probability model. Therefore, there are two
options to asses the validity of the polynomials. We will first compare
the linear output before converting it into probabilities, both in the
network and in the obtained polynomials:

\begin{CodeChunk}
\begin{CodeInput}
R> prediction_nn_linear <- predict(nn_cls, test_cls_x, verbose = FALSE)
R> prediction_poly_linear <- predict(poly_nn_cls, newdata = test_cls_x)
\end{CodeInput}
\end{CodeChunk}

\begin{CodeChunk}
\begin{figure}

{\centering \includegraphics[width=1\linewidth]{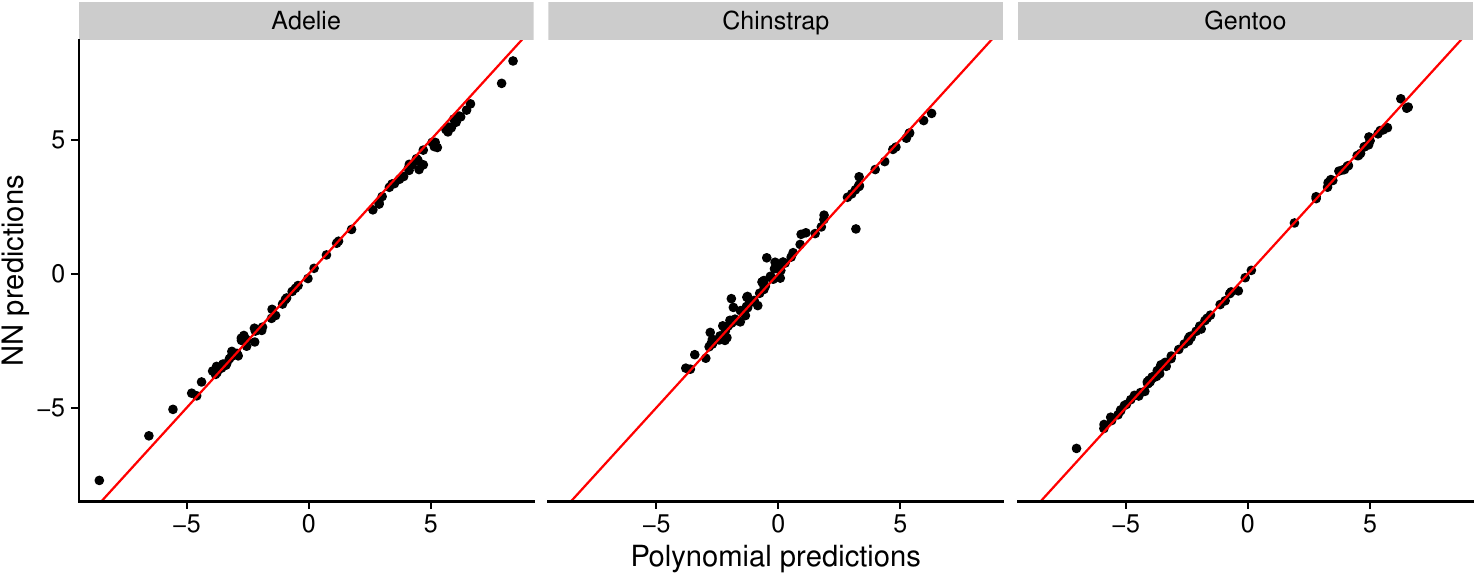} 

}

\caption[Linear predictions from the neural network model compared with
the predictions from the polynomial representation for the 3 species]{Linear predictions from the neural network model compared with
the predictions from the polynomial representation for the 3 species.}\label{fig:classification-compare}
\end{figure}
\end{CodeChunk}

Figure \ref{fig:classification-compare} showcases that the linear
predictions are quite similar between the neural network and the
obtained polynomials. To evaluate the accuracy in the obtained
classification, we need to be able to predict classes using the
polynomial. We can achieve this by applying the same probability model,
followed by a confusion matrix to compare the classes assigned by the
neural network and the polynomial. Here, we add a \emph{softmax} layer
to convert the linear output into probabilities, and then use an
\emph{argmax} function to assign the class with the highest probability.

\begin{CodeChunk}
\begin{CodeInput}
R> # Define the model probability model and then predict
R> prediction_nn_cls <- keras_model_sequential() %>%
+   nn_cls() %>%
+   layer_activation_softmax() %>%
+   layer_lambda(k_argmax) %>%
+   predict(test_cls_x, verbose = FALSE)
R> prediction_poly_cls <- keras_model_sequential() %>%
+   layer_activation_softmax() %>%
+   layer_lambda(k_argmax) %>%
+   predict(prediction_poly_linear, verbose = FALSE)
R> 
R> # NN vs original response
R> caret::confusionMatrix(
+   as.factor(prediction_nn_cls), as.factor(test_cls_y))$table
\end{CodeInput}
\begin{CodeOutput}
          Reference
Prediction  0  1  2
         0 36  0  0
         1  1 18  0
         2  0  0 28
\end{CodeOutput}
\begin{CodeInput}
R> # NN vs polynomial representation
R> caret::confusionMatrix(
+   as.factor(prediction_nn_cls), as.factor(prediction_poly_cls))$table
\end{CodeInput}
\begin{CodeOutput}
          Reference
Prediction  0  1  2
         0 36  0  0
         1  0 19  0
         2  0  0 28
\end{CodeOutput}
\end{CodeChunk}

Finally, the plot method used on the obtained \code{nn2poly} object
allows us to visualize the top-n important coefficients in each class
polynomial.

\begin{CodeChunk}
\begin{CodeInput}
R> plot(poly_nn_cls, n = 6)
\end{CodeInput}
\begin{figure}

{\centering \includegraphics[width=1\linewidth]{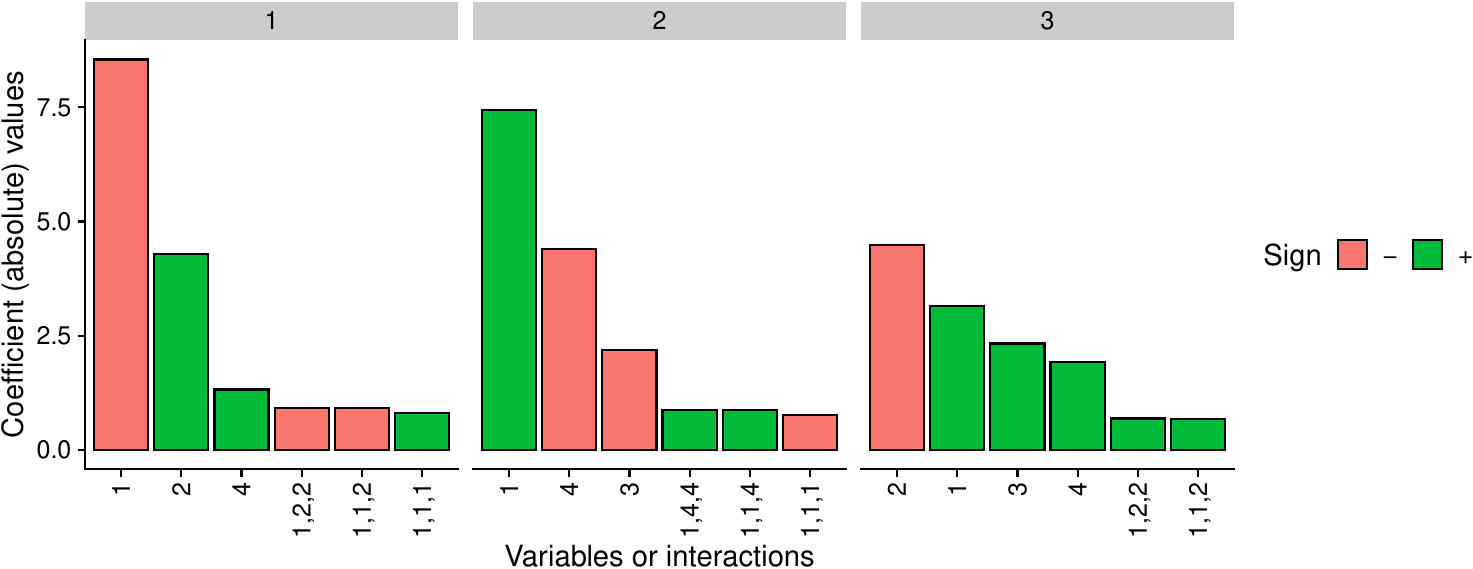} 

}

\caption[Top 6 most important coefficients from the polynomial representation for each class]{Top 6 most important coefficients from the polynomial representation for each class.}\label{fig:classification-plot}
\end{figure}
\end{CodeChunk}

In this case (see Figure \ref{fig:classification-plot}), it is clear
that the most important coefficients are assigned to single variables,
while some interactions of order 3 appear (note that, as the data is
scaled to the \([-1,1]\) interval, the coefficients of order 3 will be
multiplied by three variables lower than \(\vert1\vert\), which will
have a lower total monomial value). In the case of Adelie and Chinstrap,
variable 1 (\code{bill_length_mm}) is the highest absolute value
coefficient, with a negative effect for Adelie and positive for
Chinstrap. In the case of Gentoo, variable 2 (\code{bill_depth_mm}) is
the one with highest (negative) effect on that class. One can also
characterize each class by their coefficients: as an example, Adelie
penguins will have in general, a short bill (negative variable 1,
\code{bill_length_mm}), deeper bills (positive variable 2,
\code{bill_depth_mm}) and with a smaller effect higher weights (positive
variable 4, \code{body_mass_g}), while the \code{flipper_length_mm} does
not seem to be relevant for this species (variable 3).

\hypertarget{comparison-with-similar-frameworks}{%
\section{Comparison with similar
frameworks}\label{comparison-with-similar-frameworks}}

\label{section:comparison}

In this section, we will explore the two most complete packages in
\proglang{R} that implement neural networks interpretability solutions,
namely \pkg{Neuralsens}
\citep{pizarrosoNeuralSensSensitivityAnalysis2022} and \pkg{innsight}
\citep{koenenInnsightGetInsights2023}. As explained in Section
\ref{section:related_interpret}, \pkg{NeuralNetTools}
\citep{beckNeuralNetToolsVisualizationAnalysis2018}, is another package
in \proglang{R} devoted to neural network explanations , which
implements visualization tools and some interpretability methods such as
Olden \citep{oldenAccurateComparisonMethods2004} and Garson
\citep{garsonInterpretingNeuralnetworkConnection1991} algorithms among
others. These methods are already covered in
\citep{pizarrosoNeuralSensSensitivityAnalysis2022} and therefore will
not be explored here for comparisons, specially as the Garson algorithm
only accepts single hidden layer networks.

\subsection[Package NeuralSens]{Package \pkg{NeuralSens}}

\pkg{NeuralSens} provides neural network explanations based on
performing sensitivity analysis of neural networks using partial
derivatives of the output with respect to the input. It has been
implemented for several \proglang{R} neural network packages such as
\pkg{neuralnet}, \pkg{h2o} or \pkg{caret}, but not for the deep learning
frameworks used in \pkg{nn2poly}, i.e., \pkg{tensorflow} and
\pkg{torch}. However, non supporter frameworks can be used as input to
the main function \code{SensAnalysisMLP()} using the \code{numeric}
method, which concatenates the weights of each layer in the appropriate
order, with the bias at the first position. In particular, this can be
done for \pkg{tensorflow} sequential models following Appendix A in
\citep{pizarrosoNeuralSensSensitivityAnalysis2022}. Continuing with the
example from Section \ref{section:main_nn2poly}, we will explore the
explanations for the \code{nn_reg_const} network. Recall that the
original polynomial it was trained on was
\(Y = 2 - 2X_1 + 5X_2X_3 + 3X_4\).

\begin{CodeChunk}
\begin{CodeInput}
R> model_weights <- get_weights(nn_reg_const)
R> neural_struct <- c(nrow(model_weights[[1]]),
+                    sapply(model_weights[c(FALSE, TRUE)], nrow))
R> wts <- do.call(c, lapply(seq(2, length(model_weights), 2), function(i)
+   rbind(model_weights[[i]], model_weights[[i - 1]])))
R> # Activation functions defined manually
R> # Linear needs to be added at the start in NeuralSens notation
R> actfunc <- c("linear", "tanh", "tanh", "tanh", "linear")
\end{CodeInput}
\end{CodeChunk}

Then, with the input in the needed form, \code{SensAnalysisMLP()} can be
used to obtain the sensitivity analysis which can be summarized in two
different plots:

\begin{CodeChunk}
\begin{CodeInput}
R> library("NeuralSens")
R> 
R> sens_keras <- SensAnalysisMLP(wts, trData = data_reg,
+                               mlpstr = neural_struct,
+                               coefnames = names(data_reg)[1:5],
+                               output_name = names(data_reg)[5+1],
+                               actfunc = actfunc,
+                               plot = FALSE)
R> 
R> # Plots of sensitivity analysis results
R> plot(sens_keras)
R> plot(sens_keras, plotType = "features")
\end{CodeInput}
\begin{figure}

{\centering \includegraphics[width=0.495\linewidth]{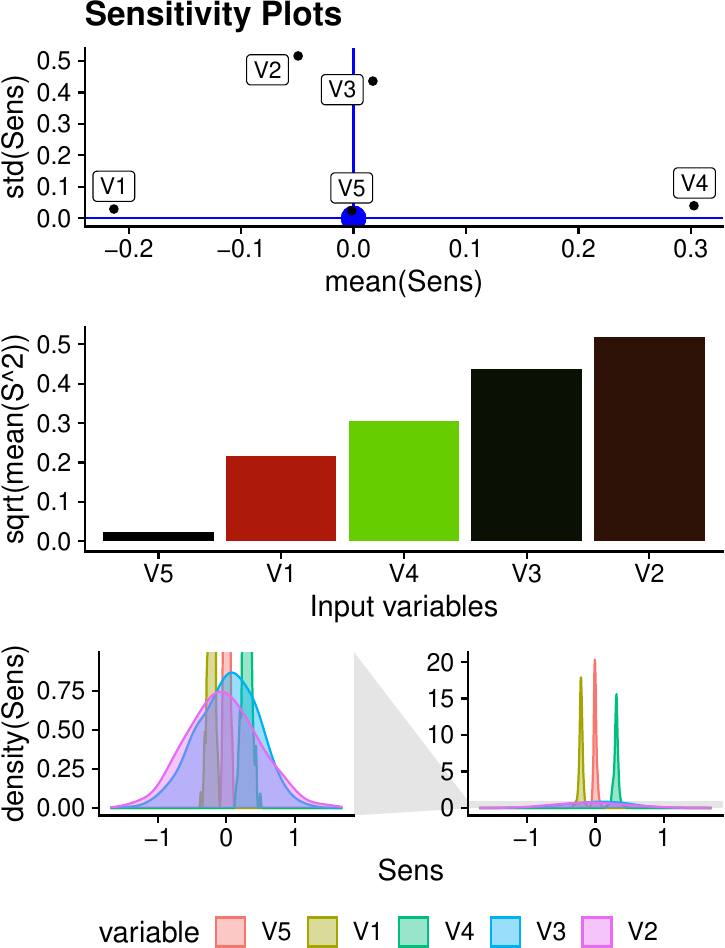} \includegraphics[width=0.495\linewidth]{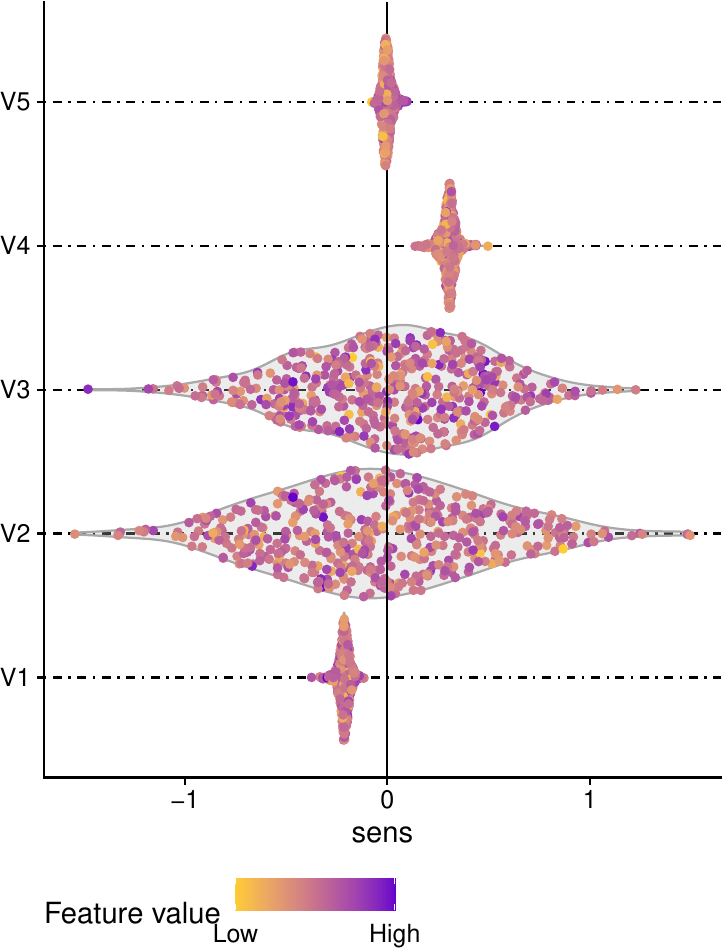} 

}

\caption[Sensitivity analysis results from \pkg{NeuralSens}]{Sensitivity analysis results from \pkg{NeuralSens}.}\label{fig:comparison-neuralsens-plot}
\end{figure}
\end{CodeChunk}

In Figure \ref{fig:comparison-neuralsens-plot}, the interaction effect
is not easy to disentangle. The effects of non-interacting variables is
well captured in the density plots (left side, bottom panels), as
indicated by the high spikes for variables 5, 1 and 4. Their magnitudes
are also accurately represented by the squared sensitivity (left side,
center panel), with V1 being clearly negative (red) and V4 positive
(green). However, the effect of the interaction between variables 2 and
3 is not easily identifiable, due to their density plots overlapping
over the whole space, as it can also be seen in the right side panel.
Furthermore, the magnitude of their effect (left side, center panel) is
high but the sign is not clear. This example with a single interaction
evidences why complex data might be difficult to explain when the
interpretability method does not account for interactions. This effect
can increase significantly when there are several interactions included
in the original data.

\subsection[Package innsight]{Package \pkg{innsight}}

Package \pkg{innsight} implements several XAI techniques specific to
neural networks and also other model agnostic methods. In this
comparison, we will focus on some of them: LRP and Gradient \(\times\)
Input as networks specific methods, and SHAP and LIME as model agnostic
methods. Note that these are local interpretability methods, but
\pkg{innsight} provides global plots that aggregate the relevances or
importance values for each variable (in absolute value) as boxplots
offering a general overview of the model behavior as a whole.

In this case, \pkg{innsight} directly supports sequential models from
\pkg{torch}, \pkg{keras} or \pkg{neuralnet}, which all of them can be
converted into \pkg{torch} models with the needed structure to be used
with their explainers.

\begin{CodeChunk}
\begin{CodeInput}
R> library("innsight")
R> 
R> # Convert the model
R> converter <- convert(nn_reg_const)
R> 
R> # Apply local methods
R> result_LRP <- LRP$new(converter, test_reg_x)
R> result_grad <- Gradient$new(converter, data = test_reg_x)
R> result_SHAP <- SHAP$new(converter, data = test_reg_x, data_ref = train_reg_x)
R> result_LIME <- run_lime(converter, test_reg_x, data_ref = train_reg_x)
R> 
R> # Plot a aggregated plot of all given data points in argument 'data' 
R> patchwork::wrap_plots(list(
+   (plot_global(result_LRP))@grobs[[1]] + ggtitle("LRP"),
+   (plot_global(result_grad))@grobs[[1]] + ggtitle("Gradient x Input"),
+   (plot_global(result_SHAP))@grobs[[1]] + ggtitle("SHAP"),
+   (plot_global(result_LIME))@grobs[[1]] + ggtitle("LIME")),
+   ncol = 2
+ )
\end{CodeInput}
\begin{figure}

{\centering \includegraphics[width=1\linewidth]{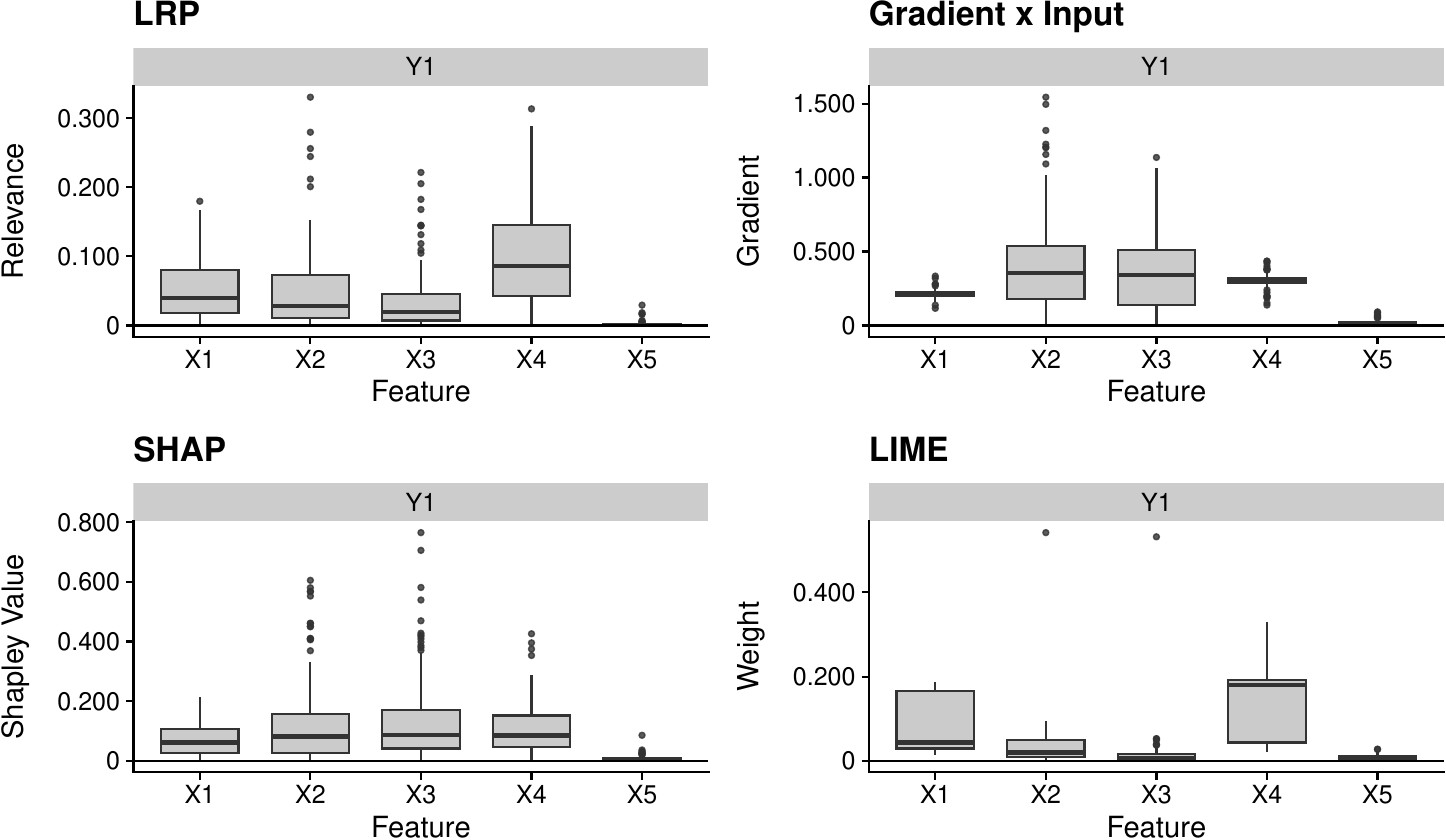} 

}

\caption[Importance plots for the 4 methods provided by \pkg{innsight}]{Importance plots for the 4 methods provided by \pkg{innsight}.}\label{fig:comparison-insight-plot}
\end{figure}
\end{CodeChunk}

In Figure \ref{fig:comparison-insight-plot}, the different methods
provide radically different results, which indicates that at least some
of them might not be well fitted to explain models trained on data with
such interactions. In particular, LIME performs poorly by assigning
really small coefficients to variables 2 and 3, the interacting ones,
while having high values and variability in the variables appearing
alone in the polynomial, i.e., 1 and 4. In the case of Gradient
\(\times\) Input, variables 2 and 3 have higher variability compared to
the other variables and to the previous linear example, which might be a
way of detecting the presence of interacting variables. In the cases of
SHAP and LRP, it does not seem easy to note that there is an interaction
appearing in the model; at most, some extreme outliers appear in the
boxplots for variables 2 and 3 in both cases. However, according to LRP,
variable 4 has a higher relevance with respect to the other variables.
In all methods, variable 5 is again correctly identified as having no
effect on the output. However, all methods lack information about the
sign of those importances in the global model, but this can be explored
locally.

\hypertarget{conclusions}{%
\section{Conclusions}\label{conclusions}}

\label{section:conc}

The \pkg{nn2poly} package presented in this paper supposes a novel
contribution to the area of ML and AI interpretability and
explainability, with the implementation of the NN2Poly method. It allows
the representation of a trained feed-forward MLP neural network as a
polynomial, that can be interpreted in terms of the obtained
coefficients, providing therefore a explanation of the model predictions
in terms of both the single variables and their interactions up to the
desired order.

This method directly supports neural networks built and trained within
the \pkg{tensorflow}+\pkg{keras} or \pkg{torch}+\pkg{luz} frameworks,
which are the two main environments in deep learning. Furthermore, a
implementation of the required constraints during neural network
training is provided for both frameworks. Other neural network models
can be used as long as their models can be represented as a list of
weight matrices with the needed activation functions naming each matrix
at each layer.

The obtained polynomial has the main use of providing interpretable
coefficients for each of the original variables as well as their
interactions, a feature that other interpretability methods do not
always provide. This is in itself an important contribution, which can
be of special interest in applied fields where explanations of the
models are enforced by administrative regulations or where informed
decision making is critical to the business. However, the polynomial
representation has several other benefits, such as offering the
possibility of inspecting the internal behavior as polynomials at each
neuron and layer can be obtained, due to the iterative approach in the
theoretical NN2Poly algorithm. It can also provide faster inference
times if the neural network predictions are replaced by the polynomial
predictions.

Future work on the \pkg{nn2poly} package may provide new functionalities
such as:

\begin{itemize}
\tightlist
\item
  Directly support the usage of \code{nn2poly()} on neural network
  classes from other packages. This may also suppose the implementation
  of the needed constraints during training, however, this may not be
  possible for all packages if their model training is not flexible
  enough.
\item
  Extend the usage of \code{nn2poly()} to other types of neural networks
  different from feed-forward MLPs, such as Convolutional Neural
  Networks (CNN) or Recurrent Neural Networks (RNN).
\item
  Support for local explanations for a given observation, instead of the
  global model. This may allow easier comparison with other
  interpretability methods of local nature.
\end{itemize}

\section*{Acknowledgments}

This research is part of the I+D+i projects PDC2022-133359,
PID2022-137243OB-I00 and TED2021-131264B-100 funded by
MCIN/AEI/10.13039/501100011033 and European Union NextGenerationEU/PRTR.
This initiative has also been partially carried out within the framework
of the Recovery, Transformation and Resilience Plan funds, financed by
the European Union (Next Generation) through the grant ANTICIPA.

\clearpage

\renewcommand\refname{References}
\bibliography{references/main.bib,references/pkgs.bib}

\end{document}